\pdfoutput=1
\documentclass[11pt]{article}

\usepackage[]{emnlp2021}

\usepackage{times}
\usepackage{latexsym}

\usepackage[T1]{fontenc}
\usepackage[utf8]{inputenc}

\usepackage{microtype}

\usepackage{microtype}      % make text margins prettier
\usepackage{graphicx}       % import graphical images
\usepackage[T1]{fontenc}    % fix font related glitches such as "|"
\usepackage{multirow}       % merge rows in tables
\usepackage{subcaption}     % create sub-tables and sub-figures
\usepackage{amssymb}        % enable \mathbb, \mathcal
\usepackage{bold-extra}     % enable \texttt{\textbf{}}
\usepackage{bm}             % enable bold font in math mode
\usepackage[hang,flushmargin]{footmisc}  % minimize footnote indentation
\usepackage{amsmath}   % for matrices
\usepackage{makecell}  % for text spanning multiple lines in table cell
\usepackage{stfloats}  % for bottom-aligned table
\usepackage{arydshln}  % hdashline

\newcommand{\LN}{\linebreak\noindent}    % to manage inline spacing
\newcolumntype{?}{!{\vrule width 2pt}}
\newcommand{\T}{\hspace*{.5cm}} 

\usepackage{multicol}

\title{StreamSide: A Fully-Customizable Open-Source Toolkit for \\ Efficient Annotation of Meaning Representations}

\author{Jinho D. Choi \\
  Computer Science \\
  Emory University \\
  Atlanta GA 30322 USA \\
  \texttt{jinho.choi@emory.edu} \\\And
  Gregor Williamson \\
  Computer Science \\
  Emory University \\
  Atlanta GA 30322 USA \\
  {\resizebox{0.47\linewidth}{!}{\texttt{gregor.jude.williamson@emory.edu}}} \\
}

\begin{document}
\maketitle

\begin{abstract}

This demonstration paper presents StreamSide, an open-source toolkit for annotating multiple kinds of meaning representations.
StreamSide supports frame-based annotation schemes e.g., Abstract Meaning Representation (AMR) and frameless annotation schemes e.g., Widely Interpretable Semantic Representation (WISeR).
Moreover, it supports both sentence-level and document-level annotation by allowing annotators to create multi-rooted graphs for input text.
It can open and automatically convert between several types of input formats including plain text, Penman notation, and its own JSON format enabling richer annotation.
It features reference frames for AMR predicate argument structures, and also concept-to-text alignment.
StreamSide is released under the Apache 2.0 license, and is completely open-source so that it can be customized to annotate enriched meaning representations in different languages (e.g., Uniform Meaning Representations).
All\LN StreamSide resources are publicly distributed\LN through our open source project at: \url{https://github.com/emorynlp/StreamSide}.

\end{abstract}

\section{Introduction}
\label{sec:introduction}

Abstract Meaning Representation (AMR) encodes the meaning of a natural language sentence (or document) as a direct acyclic graph, in which terminal nodes correspond to concepts and edges represent semantic relations between the concepts \citep{banarescu-etal-2013-abstract}.
Below is an AMR in Penman notation \citep{matthiessen1991text}:

\begin{figure}[htbp!]
\centering
\texttt{\noindent
\begin{tabular}{@{}l@{}}
(w / want-01 \\
\hphantom{(w }:ARG0 (b / boy) \\
\hphantom{(w }:ARG1 (b2 / believe-01 \\
\hphantom{(w :ARG1 (b2 }:ARG0 (g / girl) \\
\hphantom{(w :ARG1 (b2 }:ARG1 b))
\end{tabular}
}
    \caption{An abstract meaning representation for the sentence, ``\textit{the boy wants the girl to believe him}''.}
    \label{fig:amr}
\end{figure}

\noindent At the time of writing, there is one existing tool for producing AMR annotation in Penman notation; the online AMR editor of \citet{hermjakob2013amr}.\footnote{Editor: \url{https://amr.isi.edu/editor.html}}
Despite its various functionalities, and its contribution to the early-stage development of AMR, there are a few downsides to this editor.\LN
Firstly, it is not customizable except by administrators and thus has not kept up with the latest advancements in the field, as it was last updated in 2018.
Next, the way that the editor builds graphs is rigidly top-down and therefore does not give any flexibility for annotators to create sub-graphs which can later be connected.
This means that the annotator must figure out the entire structure of the final graph before they start building it.\LN
Most significantly, it is only usable as a guest without a login and, at present, logins must be requested from the administrators.
Finally, annotators need to be online in order to use the editor.

With these drawbacks in mind, we introduce StreamSide, an offline editor for annotating graph-based meaning representations which is open-source and fully customizable.
It is Python-based and so can be easily updated to reflect developments in the field as this is the preferred programming language for computational linguists without an engineering background. 
Everyone is encouraged to add further functionality to the toolkit as we plan to build a community of users, expanding the project to various domains in multiple languages.

\section{Installation}
\label{sec:installation}

StreamSide can be installed through the Python Package Index (PyPi) using the \texttt{pip} command, and is compatible with Python versions $\geq$ 3.7:

\begin{verbatim}
pip install streamside
\end{verbatim}

\noindent Detailed instructions on updating pip and installing StreamSide in a virtual environment can be found at \url{https://pypi.org/project/streamside/}.

\begin{figure*}[htbp!]
\centering
\includegraphics[width=0.9\textwidth]{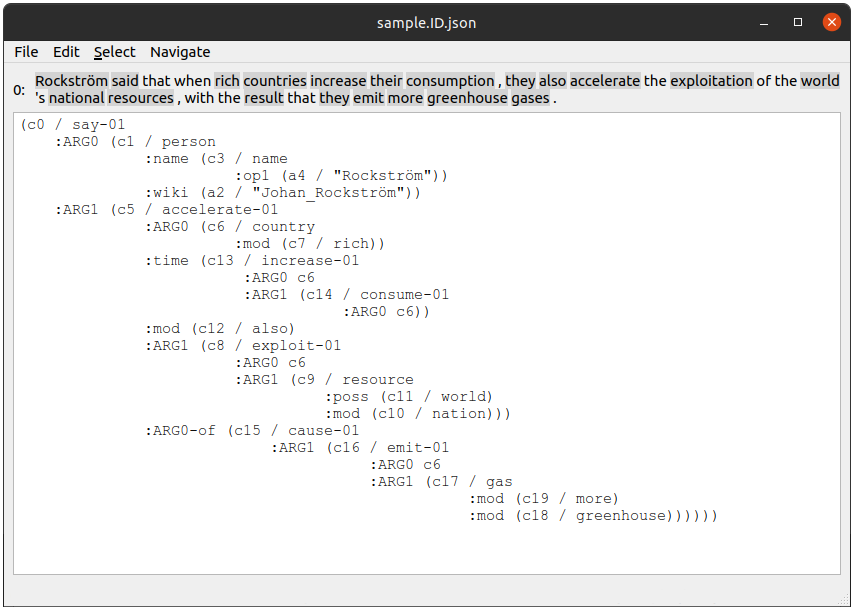}
\caption{The StreamSide GUI with a completed annotation in the workspace.}
\label{fig:GUI}
\end{figure*}

\section{Workflow}
\label{sec:workflow}

\subsection{Launching Annotation GUI}
\label{ssec:launching}

The StreamSide graph annotator is launched from the terminal.
When entering the launch command, you can specify the annotator ID (\texttt{--annotator} or \texttt{-a}), the annotation scheme (\texttt{--scheme} or \texttt{-s}), and the resource directory (\texttt{--resources} or \texttt{-r}):

\begin{verbatim}
python -m streamside.annotator
   -a ANNOTATOR_ID
  [-s ANNOTATION_SCHEME]
  [-r RESOURCE_DIR]
\end{verbatim}

\noindent The \texttt{-s} and \texttt{-r} are optional.
Currently, StreamSide has annotation support for two schemes: \texttt{amr} for Abstract Meaning Representation and \texttt{wiser} for Widely Interpretable Semantic Representation, which is a frameless representation similar to the Penman Sentence Plan Language \cite{kasper-1989-flexible}.
If \texttt{-s} is not specified, \texttt{wiser} is chosen by default.

The resource directory contains two JSON files, \texttt{concepts.json} and \texttt{relations.json}, that define concept IDs and relation labels used for the annotation, respectively.
StreamSide includes the necessary resources for English AMR and WISeR annotation.\footnote{\url{https://github.com/emorynlp/StreamSide/tree/master/streamside/resources/}}
For customizability, it also allows you to create your own resource files and specify them using the \texttt{-r} option, in which case, the \texttt{-s} option will be ignored and StreamSide will adapt to the scheme provided in the resource files.

\subsection{Opening and Saving}
\label{ssec:opening}

Annotation batches should be formatted in a plain text file, with each sentence starting on a new line.
Figure~\ref{fig:GUI} shows the StreamSide GUI, where the first sentence (starting with ID \texttt{0}) is annotated in AMR.
The annotator can navigate through sentences using the \textit{Navigate} dropdown menu or the corresponding shortcut keys that allow you to move to the previous and next sentence, or jump to a specific one.
StreamSide also has the capacity to read annotations in Penman format, which allows users to open annotations from existing AMR corpora and revise them using its comprehensive GUI.

In-progress annotation batches are automatically saved as a JSON file with the annotator ID included in the filename and can be continued or revised at a\LN later time if needed.
If a returning annotator opens a text/penman file that has been already claimed by\LN one's annotator ID, StreamSide will open the corresponding JSON file instead.
The JSON format used by StreamSide, as well as its benefits over Penman notation, are explained in Section~\ref{ssec:output}.

\begin{figure*}[htbp!]
\centering

\begin{subfigure}{0.48\textwidth}
\centering
\includegraphics[width=0.75\columnwidth]{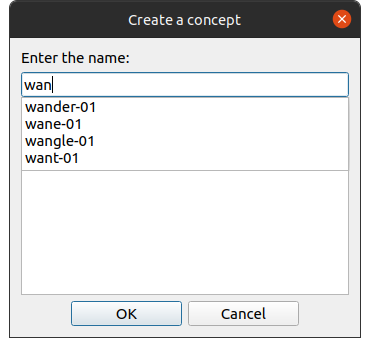}
\caption{Creating a concept with live search results.}
\label{fig:create}
\end{subfigure}
~~~~~
\begin{subfigure}{0.48\textwidth}
\centering
\includegraphics[width=0.75\columnwidth]{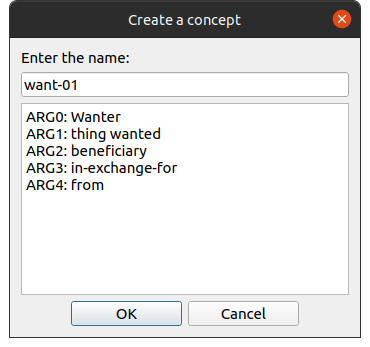}
\caption{Creating a concept with the AMR/PropBank frames.}
\label{fig:PropBank}
\end{subfigure}

\caption{Creating a concept using the Concept Dialog.}
\label{fig:create-concept}
\end{figure*}

\section{Annotation}
\label{ssec:annotating}

\subsection{Word-to-Concept Alignment}
\label{sssec:alignment}

\begin{figure}[htbp!]
\centering
\includegraphics[trim={0, 10cm, 0, 0},clip,width=\columnwidth]{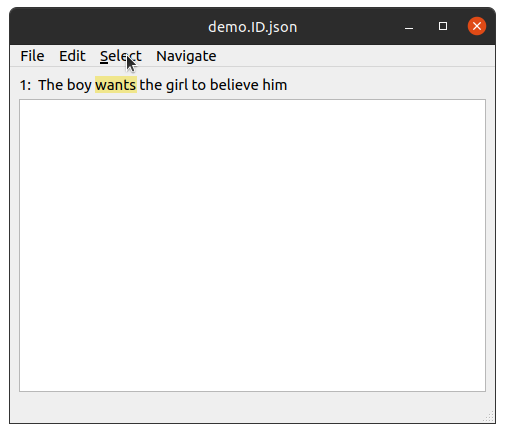}
\caption{Selecting a span of text in the GUI}
\label{fig:select}
\end{figure}

StreamSide allows you to align text spans to concepts or attributes you create.
To do this, you first select text spans using the \textit{Select} dropdown menu or pressing \texttt{x} (Figure~\ref{fig:select}).
Disjoint spans can be created by selecting the relevant sub-spans sequentially.
These are then aligned to a concept/attribute upon its creation (section~\ref{sssec:creating-conc-and-attr}).
Once a concept/attribute is created, additional text spans can be added or removed from the alignment as appropriate.
For proper alignment, input should be tokenized although alignment is optional, and annotation can be produced without it.\footnote{As text-alignment is non-trivial in certain cases (e.g., named entity, co-reference), guidelines are provided at \url{https://github.com/emorynlp/StreamSide/blob/master/docs/streamside_guidelines.md}.}

\noindent Research shows that transducer-based parsing models tend to perform better at producing AMR graphs in Penman notation when the order of arguments correspond more closely to the linear sentence order \citep{konstas-etal-2017-neural,bevilacqua-etal-2021-one}.
StreamSide orders arguments according to the order of appearances of the aligned text spans.
As with any feature of StreamSide, the default argument ordering can be easily overridden according to the needs of your annotation project.

\subsection{Creating Concepts and Attributes}
\label{sssec:creating-conc-and-attr}

Using the \textit{Edit} menu, annotators can create a concept, attribute, or relation.
When creating a concept, a popup appears featuring a search bar, which is filled with the selected span of text (Figure~\ref{fig:create}).
The live search results may consist of concepts which are used by AMR, including predicate senses from PropBank \citep{palmer-etal-2005-proposition, bonial-etal-2014-propbank} and standard concepts such as \texttt{amr-unknown}.
If a predicate sense associated with an AMR frame is chosen (e.g., \texttt{want-01}), then its argument structure will appear in the description box, as shown in Figure~\ref{fig:PropBank}.
Annotators are thus able to verify that the selected sense is the desired one.

Constants such as numbers (e.g., \texttt{23}), names\LN (e.g., \texttt{"California"}), politeness and polarity attributes (i.e., \texttt{+/-}), and mode attributes for sentential mood (i.e., \texttt{expressive} and \texttt{imperative}) are created using \textit{Create Attribute} under the \textit{Edit} menu.
Attributes are displayed in the StreamSide workspace with a variable ID and parentheses so that they can be aligned with the text and interact with other nodes, be deleted, or be updated.
When the output of StreamSide is converted to Penman notation, however, attributes will appear without parentheses or a variable ID.

\begin{figure*}[htbp!]
\centering

\begin{subfigure}[t]{0.48\textwidth}
\centering
\includegraphics[width=0.8\columnwidth]{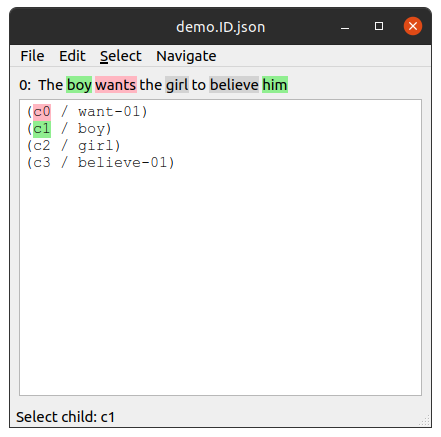}
\caption{The workspace populated with concepts. The parent node and its aligned text are highlighted in red, and the child node and its aligned text are in green.}
\label{fig:concepts-highlighted}
\end{subfigure}
~~~~~
\begin{subfigure}[t]{0.48\textwidth}
\centering
\includegraphics[width=\columnwidth]{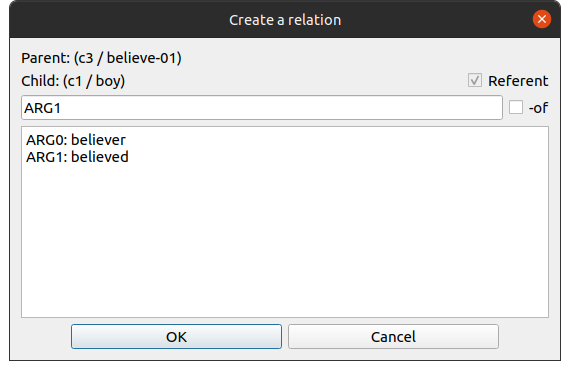}
\caption{Creating a relation with the \textit{referent} box automatically checked in the Relation Dialog.}
\label{fig:ref-and-role}
\end{subfigure}

\caption{Creating a relation using the main workspace and the Relation Dialog.}
\label{fig:create-relation}
\end{figure*}

An advantage of StreamSide over existing editors such as the online AMR editor \citep{hermjakob2013amr} is the degree of flexibility annotators have when creating graphs.
The original AMR editor builds graphs in a strictly top-down fashion, while StreamSide allows for both top-down and bottom-up graph construction, as well as a combination of the two.
For instance, the annotator can populate the workspace with the concepts that will be used prior to creating the relations between them (see Figure~\ref{fig:concepts-highlighted}).
Moreover, since StreamSide workspace can accommodate more than one root, annotators can build multiple sub-graphs before combining them into the final graph.
This amount of freedom when figuring out how to build a graph is particularly useful if the graph has a complex structure.

\subsection{Creating Relations}

To create a relation between two nodes, annotators first highlight and select the parent node using the \textit{Select} menu before selecting the child node.
This will leave the parent node and its associated span of text highlighted in red, and the child and its associated span in green, as shown in Figure~\ref{fig:concepts-highlighted}.

The annotator then creates a relation using the \textit{Edit} menu.
If the parent node is associated with an AMR frame, the corresponding argument structure will be displayed in the description box.
When creating a relation, two check boxes for handling reentrancies are available to annotators (Figure~\ref{fig:ref-and-role}).
The first is the \textit{referent} box which should be checked for instances of co-reference.
This box will be checked automatically if a concept is already in a relation with another.
The second check box is the \textit{-of} box which creates an inverse relation.

\subsection{Deleting and Updating}

All created concepts, attributes, and relations can be deleted or updated using the \textit{Edit} menu.
For concepts and attributes, first highlight their IDs and choose an appropriate action from the menu.\LN
For relations, highlight their labels instead and then choose an action.

\subsection{Keyboard shortcuts}
\label{ssec:shortcuts}

Although StreamSide can be used in a point-and-click fashion, annotators are encouraged to use the available keyboard shortcuts to increase annotation speed.
Shortcuts are included for opening/saving files, navigating sentences, selecting spans and nodes, creating, deleting, or updating concepts, attributes, or relations, as well as checking the referent and inverse boxes when creating relations.\LN
When keyboard shortcuts are utilized, the use of mouse is only needed for highlighting text spans.

\subsection{Annotation Output}
\label{ssec:output}

StreamSide saves annotation files in a JSON format.\LN
Figure~\ref{fig:json} shows the AMR graph from Figures~\ref{fig:amr} in our JSON format.
These JSON files can be converted into Penman files using StreamSide's built-in JSON-to-Penman converter, where \texttt{IN\_PATH} is the path to a directory containing input JSON files and \texttt{OUT\_PATH} is the directory path to save output Penman files (optional):

\begin{verbatim}
python
  -m streamside.json_to_penman
  -i IN_PATH [-o OUT_PATH]
\end{verbatim}

\noindent The alignment information is also represented in Penman files created using the JSON-to-Penman converter.
However, the alignment notation used is different from that in the AMR 3.0 release \citep{knight:2020a}, and is designed to be more intuitive.
Once a sufficiently large corpus of aligned annotations is produced, models can be trained on aligned data for improved parsing results.

\begin{figure}[!ht]
\vspace{-0.2cm}
\resizebox{\columnwidth}{!}{\texttt{\noindent
\begin{tabular}{@{}l@{}}
\{\vspace*{-0.1cm}
\\ \T  "graphs":[\{
\\ \T\T"tid":"demo.1",
\\ \T \T    "annotator":"ID",
\\ \T \T    "last\_saved":"04/17/2021 11:23:42",
\\  \T \T   "tokens":["The", "boy", "wants", "the",  
\\ \T\T \T\T\T\T\hspace{0.3cm}"girl", "to", "believe", "him"],
\\  \T \T   "concepts":\{
\\ \T \T \T "c0":\{
\\ \T \T \T \T "name":"want-01", 
\\ \T \T \T \T "token\_ids":[2],
\\ \T \T \T \T"attribute":false,
\\ \T \T \T \T "first\_token\_id":2\}, 
\\ \T \T \T "c1":\{
\\ \T \T \T \T"name":"boy", 
\\ \T \T \T \T"token\_ids":[1, 7], 
\\ \T \T \T \T"attribute":false, 
\\ \T \T \T \T"first\_token\_id":1\},  
\\ \T \T \T "c2":\{
\\ \T \T \T \T"name":"girl",
\\ \T \T \T \T"token\_ids":[4],
\\ \T \T \T \T"attribute":false,
\\ \T \T \T \T"first\_token\_id":4\}, 
\\ \T \T \T "c3":\{
\\ \T \T \T \T"name":"believe-01", 
\\ \T \T \T \T"token\_ids":[6], 
\\ \T \T \T \T"attribute":false, 
\\ \T \T \T \T"first\_token\_id":6\} \vspace*{-0.1cm}
\\ \T \T \}, 
\\ \T \T  "relations":\{
\\ \T \T \T "r0":\{
\\ \T \T \T \T"parent\_id":"c0", 
\\ \T \T \T \T"child\_id":"c1", 
\\ \T \T \T \T"label":"ARG0", 
\\ \T \T \T \T"referent":false\}, 
\\ \T \T \T "r1":\{
\\ \T \T \T \T"parent\_id":"c0", 
\\ \T \T \T \T"child\_id":"c3", 
\\ \T \T \T \T"label":"ARG1", 
\\ \T \T \T \T"referent":false\}, 
\\ \T \T \T "r2":\{
\\ \T \T \T \T"parent\_id":"c3", 
\\ \T \T \T \T"child\_id":"c2", 
\\ \T \T \T \T"label":"ARG0", 
\\ \T \T \T \T"referent":false\}, 
\\ \T \T \T "r3":\{
\\ \T \T \T \T"parent\_id":"c3", 
\\ \T \T \T \T"child\_id":"c1", 
\\ \T \T \T \T"label":"ARG1", 
\\ \T \T \T \T"referent":true\} \vspace*{-0.1cm}
\\ \T \T \},
\\  \T \T   "covered\_token\_ids":[1, 2, 4, 6, 7],
\\  \T \T   "\_concept\_id":4,
\\  \T \T "\_relation\_id":4 \vspace*{-0.1cm}
\\  \T\}]\vspace*{-0.1cm}
\\ \}
\end{tabular}}
}
\caption{The JSON format of the StreamSide annotation for the AMR annotation in Figure~\ref{fig:amr}.}
\label{fig:json}
\end{figure}

\noindent There are several advantages to storing graphs in the JSON format.
Firstly, it is straightforwardly machine-readable, and does not require familiarity\LN with additional APIs.
Secondly, this format is capable of saving more meta information than the Penman notation, and it contains annotation-to-text alignment for attributes.
This is particularly important for the negative particle \textit{not} which is often aligned with the polarity attribute `\texttt{-}'.
In Penman notation, this alignment information is lost since the attribute is not associated with a variable name.
Finally, although graph similarity can be evaluated for Penman graphs using the \textit{Smatch} metric \citep{cai-knight-2013-smatch}, the JSON format is already stored as lists of triples which means that evaluating features of graph similarity beyond those traditionally considered by Smatch can be done easily.
Likewise, researchers can quickly produce a fine-grained Smatch breakdown like that proposed by \citet{damonte-etal-2017-incremental} by isolating the F-score for specific (groups of) roles, concepts, or attributes.

Finally, StreamSide can convert from Penman to JSON using the following command, where \texttt{IN\_PATH} is the path to a directory containing input Penman files and \texttt{OUT\_PATH} is the directory path to save output JSON files (optional):

\begin{verbatim}
python
  -m streamside.penman_to_json
  -i IN_PATH [-o OUT_PATH]
\end{verbatim}

\noindent This allows annotators to edit existing annotations and add alignment data to existing corpora using the StreamSide GUI.

\section{Customization}
\label{sec:customization}

StreamSide is open source and released under the Apache 2.0 license, which gives a freedom for developers to customize it for their own projects.
As such, future improvements and additional functionality can be updated collaboratively by researchers and annotators.

The StreamSide directory consists of a \textit{resources} folder in which frames, roles, and descriptions to aid annotation are stored in JSON formats.
These can be modified as needed.
This has the advantage that researchers can add novel annotation schemes to StreamSide as they are developed.
For instance, UMR \citep[Unfiform Meaning Representation;][]{van2021designing} features document-level representations for cross sentential dependencies.
Because StreamSide can accommodate multi-rooted graphs,\LN it can be used to annotate more than one sentences simultaneously in addition to a document level representation.
All resources including the source codes for StreamSide are publicly available at \url{https://github.com/emorynlp/StreamSide}.

\section{Annotation Statistics}
\label{sec:annotation-stats}

The StreamSide annotation tool has been used to annotate a dialogue corpus of 1,000 sentences from a number of sources including EmpatheticDialogues \cite{DBLP:journals/corr/abs-1811-00207}, DailyDialog \cite{DBLP:journals/corr/abs-1710-03957}, Boston English Centre\footnote{900 English Conversational Sentences from Boston English Centre: \url{https://youtu.be/JP5LYRTZtjw}}, and PersonaChat \cite{gu2020dually}.
The annotations were completed in both AMR and WISeR and are available at the GitHub repository.\footnote{At the time these corpora were produced text-to-annotation alignment was not standardized across annotators. However, we are working on standardizing the alignment data post hoc.}
The average sentence length of this corpus is 8.35 tokens.
First-time annotators ($n = \text{6}$) were timed when annotating a subset of these sentences. 
The average annotation time per sentence is 131 seconds in AMR and 126 seconds in WISeR, demonstrating that StreamSide is an effective annotation tool for producing rapid, gold standard annotations.
\section{Conclusions}
\label{sec:conclusions}

This paper presents StreamSide, an open source toolkit for annotating meaning representations.
This toolkit has a number of features which make it ideal for annotation projects: it can build graphs flexibly, supporting both top-down and bottom-up construction; it supports optional annotation-to-text alignment; it can annotate at the sentence-level as well as the document-level; and it can read and write files in a number of standard formats.
Most importantly, StreamSide is completely open source.\LN
As such, we hope that researchers and annotators may collaborate to develop this toolkit both in terms of annotation capability and functionality.

\section*{Acknowledgements}

We gratefully acknowledge the support from the ETRI (Electronics and Telecommunications Research Institute).\footnote{ ETRI: \url{https://www.etri.re.kr}}
Any opinions, findings, and conclusions or recommendations expressed in this material are those of the authors and do not necessarily reflect the views of ETRI.

% Entries for the entire Anthology, followed by custom entries
\bibliography{emnlp2021}
\bibliographystyle{acl_natbib}

\end{document}